 \documentclass[11pt]{article}
 \usepackage{amsmath,amsthm,amsfonts,amssymb}  
\usepackage[margin=1in]{geometry}

\usepackage{enumerate,wrapfig,placeins}

\newtheorem{theorem}{Theorem}[section]
\newtheorem{lemma}[theorem]{Lemma}
\newtheorem{proposition}[theorem]{Proposition}

\usepackage{algorithm}
\usepackage{algpseudocode}

\newlength{\noteWidth}
\long\def\notes#1{\ifinner
	{\tiny #1}
\else
\marginpar{\parbox[t]{\noteWidth}{\raggedright\tiny #1}}
\fi}
\def\notes#1{}

\usepackage{textgreek,upgreek,bm}

\usepackage{titlesec}

\usepackage[usenames, dvipsnames]{color}

\usepackage[pdfa,hidelinks]{hyperref}

\usepackage{optidef}
\usepackage{graphics} 
\usepackage{subfigure}
\usepackage{amsmath} 
\usepackage{amssymb}  
\usepackage{balance}

\usepackage{verbatim}

\graphicspath{{figs/}}

\usepackage{cleveref}

\Crefname{corollary}{Corollary}{Corollaries}
\Crefname{eqnarray}{eq.}{eqs.}
\Crefname{equation}{eq.}{eqs.}

\Crefname{figure}{Fig.}{Figs.}
\Crefname{tabular}{Tab.}{Tabs.}
\Crefname{table}{Tab.}{Tabs.}
\Crefname{proposition}{Prop.}{Propositions}
\Crefname{theorem}{Thm.}{Thms.}
\Crefname{definition}{Def.}{Defs.} 
\Crefname{section}{Section}{Sections}
\Crefname{lemma}{Lemma}{Lemmas}
\Crefname{assumption}{Assumption}{Assumptions}

\def\urls#1{{\footnotesize\url{#1}}}

\def\mindex#1{\index{#1}}

\DeclareFontFamily{U}{mathx}{\hyphenchar\font45}
\DeclareFontShape{U}{mathx}{m}{n}{<-> mathx10}{}
\DeclareSymbolFont{mathx}{U}{mathx}{m}{n}
\DeclareMathAccent{\widebar}{0}{mathx}{"73}

\def\barUpupsilon{\widebar{\Upupsilon}}

\makeatletter
\newcommand\gobblepars{%
    \@ifnextchar\par%
        {\expandafter\gobblepars\@gobble}%
{}}
\makeatother

\def\whamrm#1{\smallbreak\pagebreak[3]%
	\noindent\text{\rm#1}\ \ \gobblepars}

\def\wham#1{\smallbreak\pagebreak[3]%
	\noindent\textup{\textbf{#1}}\ \ \gobblepars}

\def\Tdiff{\mathcal{D}}

\def\odestate{\upvartheta}

\def\fee{\upphi}
\def\feex{\widetilde{\fee}}

\def\occm{\upvarpi}

\def\elig{\zeta}

\def\uQ{\underline{Q}}

\def\disc{\gamma}

\def\tmptheta{\theta}

\newcommand{\bbblot}{\raise1pt\hbox{\vrule height .4ex width .4ex depth .05ex}}

\long\def\beginbox#1\endbox{\subsection*{}%
\hbox{\hspace{.05\hsize}\defbox{\medskip#1\bigskip}}%
\subsection*{}}

\def\endbox{}

 \def\archival#1{} 

\def\FRAC#1#2#3{\genfrac{}{}{}{#1}{#2}{#3}}

\def\ddt{{\mathchoice{\FRAC{1}{d}{dt}}%
{\FRAC{1}{d}{dt}}%
{\FRAC{3}{d}{dt}}%
{\FRAC{3}{d}{dt}}}}

\def\ddtp{{\mathchoice{\FRAC{1}{d^{\hbox to 2pt{\rm\tiny +\hss}}}{dt}}%
{\FRAC{1}{d^{\hbox to 2pt{\rm\tiny +\hss}}}{dt}}%
{\FRAC{3}{d^{\hbox to 2pt{\rm\tiny +\hss}}}{dt}}%
{\FRAC{3}{d^{\hbox to 2pt{\rm\tiny +\hss}}}{dt}}}}

\def\ddyp{{\mathchoice{\FRAC{1}{d^{\hbox to 2pt{\rm\tiny +\hss}}}{dy}}%
{\FRAC{1}{d^{\hbox to 2pt{\rm\tiny +\hss}}}{dy}}%
{\FRAC{3}{d^{\hbox to 2pt{\rm\tiny +\hss}}}{dy}}%
{\FRAC{3}{d^{\hbox to 2pt{\rm\tiny +\hss}}}{dy}}}}

\def\half{{\mathchoice{\FRAC{1}{1}{2}}%
{\FRAC{1}{1}{2}}%
{\FRAC{3}{1}{2}}%
{\FRAC{3}{1}{2}}}}

\def\state{{\sf X}}

\def\ustate{{\sf U}} 
 
\def\zstate{{\sf Z}}

\def\bfmath#1{{\mathchoice{\mbox{\boldmath$#1$}}%
{\mbox{\boldmath$#1$}}%
{\mbox{\boldmath$\scriptstyle#1$}}%
{\mbox{\boldmath$\scriptscriptstyle#1$}}}}

\def\bfPhi{\bfmath{\Phi}}
\def\bfPsi{\bfmath{\Psi}}

\def\bfmU{\bfmath{U}}

\def\bfmX{\bfmath{X}}

\def\bfmY{\bfmath{Y}}

\def\bfmhhaY{\bfmath{\hhaY}} 
\def\bfmhhaY{\hbox to 0pt{$\widehat{\bfmY}$\hss}\widehat{\phantom{\raise 1.25pt\hbox{$\bfmY$}}}}

\def\bfmZ{\bfmath{Z}}

\def\haA{\widehat A}

\def\tiltheta{{\tilde \theta}}

\def\tilg{\tilde g}

\def\clE{{\cal E}}

\def\clZ{{\cal Z}}

\def\eqdef{\mathbin{:=}}

\def\Prob{{\sf P}}

\def\Expect{{\sf E}}

 \def\epsy{\varepsilon}

\def\varble{\,\cdot\,}

\def\formtmp#1#2{{\vskip12pt\noindent\fboxsep=0pt\colorbox{#1}{\vbox{\vskip3pt\hbox to \textwidth{\hskip3pt\vbox{\raggedright\noindent\textbf{#2\vphantom{Qy}}}\hfill}\vspace*{3pt}}}\par\vskip2pt%
\noindent\kern0pt}}

\def\barb{{\overline {b}}}

\def\barf{{\widebar{f}}}

\def\barA{{\bar{A}}}

\def\barpsi{{\bar{\psi}}}

{\end{list}}

\def\ass(#1:#2){(#1\ref{#1:#2})}

\def\ritem#1{
\item[{\sf \ass(\current_model:#1)}]
}

\newenvironment{recall-ass}[1]{%
\begin{description}
\def\current_model{#1}}{
\end{description}
}

 \newcommand{\blot}{\vrule height 1.1ex width .9ex depth -.1ex }
\def\qedb{\ifmmode\blot\else{\vspace{-.2cm}\unskip\nobreak\hfil
\penalty50\hskip1em\null\nobreak\hfil\blot
\parfillskip=0pt\finalhyphendemerits=0\endgraf}\fi}

  \makeatletter
\DeclareRobustCommand{\sqcdot}{\mathbin{\mathpalette\morphic@sqcdot\relax}}
\newcommand{\morphic@sqcdot}[2]{%
  \sbox\z@{$\m@th#1\centerdot$}%
  \ht\z@=.33333\ht\z@
  \vcenter{\box\z@}%
}
\makeatother

\def\qedsymbol{\hbox{\tiny$\blacksquare$}}  %

\def\qed{\ifmmode\qedsymbol\else{\unskip\nobreak\hfil
\penalty50\hskip1em\null\nobreak\hfil\qedsymbol
\parfillskip=0pt\finalhyphendemerits=0\endgraf}\fi}

\newcounter{rmnum}

\newcounter{anum}

\newcommand{\field}[1]{\mathbb{#1}}

\def\Re{\field{R}} 

\def\intgr{\field{Z}}

\def\Co{\field{C}}

\def\Prob{{\sf P}}

\def\Expect{{\sf E}}

\def\transpose{{\intercal}}

\def\trace{\hbox{\rm trace\,}}  
\def\epsy{\varepsilon}
\def\varble{\,\cdot\,}

\def\haY{\widehat{Y}}

\def\hhaY{\hbox to 0pt{$\haY$\hss}\widehat{\phantom{\raise 1.25pt\hbox{Y}}}}

\def\haA{\widehat A}

\def\haY{\widehat Y}

\def\bfPhi{\bfmath{\Phi}}

\newlength{\dhatheight}

\def\barUpupsilon{\widebar{\Upupsilon}}

\def\barphi{\bar{\phi}}

\def\cond{\text{\rm cond}}
 
\def\delrel{\delta_{\text{\sf r}}}
\def\nsnap{n_{\text{\sf s}}}

\def\psisub#1{\psi_{(#1)}}

\def\csub#1{c_{#1}}

\def\Qstar{Q^\star}
\def\uQstar{\uQ^\star}

\def\whamb{\wham{$\bullet$} }

\def\thetaPR{\theta^{\text{\tiny\sf  PR}}}
\def\thetaPR{\theta^{\text{\tiny\sf  PR}}}

\def\barthetaPR{\bar{\theta}^{\text{\tiny\sf  PR}}}

\def\tTheta{{\text{\tiny$\Theta$}}}

\def\SigmaTheta{\Sigma_\tTheta}

\def\thetaPR{\theta^{\text{\tiny\sf  PR}}}

\usepackage[usenames,dvipsnames]{color}
\usepackage{color}
\definecolor{programcode}{gray}{0.9}

\definecolor{lightgray}{gray}{0.7}

\definecolor{MyDarkBlue}{cmyk}{0.5,0.1,0,0.9}

\DeclareMathAccent{\widecheck}{0}{mathx}{"71}

\def\wham#1{\smallbreak\pagebreak[3]%
\noindent\textbf{#1}\ \ \gobblepars}

\def\rd#1{{\color{red}#1}}

\makeatletter
\def\thanks#1{\protected@xdef\@thanks{\@thanks
	\protect\footnotetext{#1}}}
\makeatother

\title{Stability and Sensitivity Analysis   of 
\\
Relative Temporal-Difference Learning: Extended Version}

\author{Masoud S. Sakha${}^\dagger$, Rushikesh Kamalapurkar${}^\dagger$ and Sean~Meyn*
\thanks{${}^\dagger$University of Florida, Department of Mechanical  and Aerospace Engineering (e-mails:  masoud.sakha@ufl.edu, rkamalapurkar@ufl.edu). }
\thanks{*University of Florida, Department of Electrical and Computer Engineering (e-mail:  meyn@ece.ufl.edu). 
Financial support from ARO award W911NF2010055, NSF award CCF~2306023, and AFRL award FA8651-24-1-0019 
		is gratefully acknowledged.  }%
}

\begin{document}

\thispagestyle{empty}

\maketitle

\begin{abstract}

Relative temporal-difference (RTD) learning was introduced to mitigate the slow convergence of TD methods when the discount factor approaches one by subtracting a baseline from the temporal-difference update. While this idea has been studied in the tabular setting, stability guarantees with function approximation remain poorly understood. This paper analyzes RTD learning with linear function approximation. We establish stability conditions for the algorithm and show that the choice of baseline distribution plays a central role. In particular, when the baseline is chosen as the empirical distribution of the state–action process, the algorithm is stable for any non-negative baseline weight and any discount factor. We also provide a sensitivity analysis of the resulting parameter estimates, characterizing both asymptotic bias and covariance. The asymptotic covariance and asymptotic bias are shown to remain uniformly bounded as the discount factor approaches one.
 
 \bigskip
 
 Extended version of manuscript submitted to the 2026 IEEE Conference on Decision and Control.

	\wham{Keywords:  reinforcement learning;  optimal control; temporal-difference learning. }
\end{abstract}
	
\thispagestyle{empty}

\def\betag{\beta^\blacklozenge}
\def\thetag{\theta^\blacklozenge}
 \def\cg{c_\blacklozenge}
 \def\Qg{Q^\blacklozenge}
 \def\Ag{A^\blacklozenge}
 \def\feeg{\fee^\blacklozenge}
 \def\upvarpig{\upvarpi^\blacklozenge}
 
\def\sft{\textsf{\footnotesize t}}
\def\thetat{\theta^{\sft}}
 \def\ct{c_{\sft}}
 \def\Qt{Q^{\sft}}
 \def\feet{\fee^{\sft}}
 \def\barft{\barf_{\sft}}

\clearpage

   \tableofcontents

\section{Introduction}
\label{s:intro}

Consider a Markov Decision Process (MDP)  with state process $\bfmX=\{ X_k : k\ge 0\} $ evolving on a state space $\state$, and input process $\bfmU=\{ U_k : k\ge 0\} $ evolving on $\ustate$.   In most of the technical results it is assumed that $\state$ and $\ustate $ are finite to avoid discussion of measurability and other technicalities.  
 
 \begin{subequations}%
 
In this paper, we consider approximation of the solution to the discounted-cost optimal control problem. For a given
discount factor $\disc\in(0,1)$ and a cost function $c\colon\state\times\ustate \to \Re$, the state-action value function is denoted by $\Qstar\colon\state\times\ustate \to \Re$, and is defined by
\begin{equation}
\Qstar(x,u) = \min \sum_{k=0}^\infty \disc^k \Expect [ c(Z_k) \mid   Z_0    = (x,u ) ]    
\label{e:Q}
\end{equation}
where   $Z_k = (X_k,U_k)$, and the minimum is over all history dependent input sequences. 
This is the \textit{Q-function} of Q-learning, which solves the Bellman equation, 
\begin{equation}
\Qstar(x,u) = c(x,u) +  \disc \Expect [ \uQstar(X_1) \mid   Z_0    = (x,u ) ]  
\label{e:BE}
\end{equation}
where $\uQstar(x) \eqdef\min_u \Qstar(x,u)$.   

\label{e:QstarAndQ}
\end{subequations}

 \begin{subequations}%

One approach to approximating $\Qstar$ is through approximate policy iteration, in which case we require approximation of a fixed-policy Q-function.   Let $\feex$ denote a randomized stationary policy, defined so that   $\Prob\{ U_k = u \mid  X_0,\dots, X_k \}  =  \feex(u \mid x)$ when $X_k = x$.  
For the fixed policy $\feex$, the discounted cost and Bellman equation take on a form similar to \eqref{e:QstarAndQ}:
\begin{align}
Q (x,u) & =   \sum_{k=0}^\infty \disc^k \Expect [ c(Z_k) \mid Z_0    = (x,u )] 
\label{e:Qfee}
\\
Q(x,u) &= c(x,u) +  \disc \Expect [ \uQ(X_1) \mid   Z_0    = (x,u ) ]  
\label{e:BEfee}
\end{align}
where the underbar has a new interpretation:    $\uQ(x) = \sum_u Q(x,u) \feex(u\mid x)$,  and in \eqref{e:Qfee}  the state-action sequence $\bfmZ$ is obtained using the policy $\feex$.

\label{e:QstarAndQfee}
\end{subequations}

In this paper we consider linear function approximation, in which $Q^\theta(x,u) = \theta^\transpose \psi(x,u)$ for a parameter vector $\theta \in \Re^d$,
and basis vector $\psi\colon\state\times\ustate\to\Re^d$.  In this setting, algorithms to estimate $Q^\star$, or the fixed policy Q-function $Q$, can be interpreted as stochastic approximation (SA) algorithms with associated mean flows of the form
\begin{equation}
\ddt \odestate_t =  \barf(\odestate_t) 
\label{e:meanflowQ}
\end{equation}
Stability of the   algorithm follows from global asymptotic stability of the mean flow.  

When convergence holds,  so that $\barf$ has a unique root $\theta^*$, finer properties such as rates of convergence are governed by the Jacobian    $\barA = \partial \barf\, (\theta^*)$.   In particular, the celebrated optimal \textit{asymptotic covariance} in the Central Limit Theorem (CLT) for SA
 is given by
\begin{equation}
		\SigmaTheta^* = \barA^{-1}  \Sigma_\Delta [{\barA}^\transpose]^{-1}
		\label{e:SigmaPRopt}
\end{equation}
in which $\Sigma_\Delta$ is the $d\times d$ ``noise covariance''---further discussion and history may be found below.      The more recent theory of \textit{asymptotic bias} also involves the inverse of this Jacobian:  
\begin{equation}
 \upbeta_\uptheta =  
 \tfrac{1}{1-\rho} \barA^{-1} \barUpupsilon  \label{e:biasformula}
\end{equation}
with $\barUpupsilon  \in\Re^d$;  
see \Cref{t:bias} for details.

A fundamental challenge arises when the discount factor $\disc$ approaches one. In this regime, the matrix $\barA$ is typically poorly conditioned, and 
an eigenvalue may approach zero as $\disc\uparrow 1$.    As a consequence, the asymptotic covariance and bias can grow large, and convergence may become arbitrarily slow. This behavior is reflected in both classical SA bounds and recent analyses of TD and Q-learning, where error bounds depend explicitly on the inverse $ \barA^{-1}$  associated with the linearized dynamics \cite{devmey17b,devmey22,CSRL}.
\notes{April 1: removed or Lyapunov solutions since this is made far clearer in lit review}

Relative temporal-difference (RTD) methods were introduced to address this issue by subtracting a baseline term from the update. These methods were originally proposed in the context of Q-learning, where they were shown empirically to improve convergence, particularly in regimes with large discount factors.   There is complete theory only for tabular Q-learning---understanding of relative methods with function approximation remains limited.

The present paper provides a systematic analysis of RTD learning with linear function approximation. Our main contribution is to show that a natural choice of baseline---based on the stationary distribution of the state–action process---yields a mean flow whose linearization remains uniformly Hurwitz and well-conditioned over $\disc \in (0,1)$. As a consequence, both asymptotic covariance and bias remain uniformly bounded. We also provide sensitivity analysis with respect to the baseline parameter.

These results connect directly with modern stochastic approximation theory, where both finite-time bounds and asymptotic covariance depend explicitly on the conditioning of the linearized mean dynamics. In this sense, RTD learning can be interpreted as a mechanism for regularizing the mean flow.

\subsection{Temporal difference methods}

\begin{subequations}%

The reason for the common notation in \eqref{e:BE}  and   \eqref{e:BEfee}  is for common notation in the presentation of algorithms, which take the following form:   
For initialization $\theta_0\in\Re^d  $, the sequence of estimates is defined recursively as
\begin{align}
\theta_{n+1} &= \theta_n + \alpha_{n+1}   \Tdiff_{n+1} \elig_n  
\label{e:Qlambda}
\\
\Tdiff_{n+1} & =  c(Z_n)   + \disc     \uQ^{\theta_n} (X_{n+1})  - Q^{\theta_n}(Z_n) 
\label{e:TD_theta_n}
\end{align} 
in which, $\{\alpha_n\}$ is a non-negative step-size sequence,     $\{\Tdiff_{n+1} \} $ is known as the temporal difference sequence, and the term  
$\uQ^{\theta_n}(x)$ is defined according to the context; see preceding discussion and  \eqref{e:uQintro}.
The $d$-dimensional  vectors in the sequence $\{ \elig_n   \}$  are known as the \textit{eligibility vectors}.  Under our assumption of linear function approximation,  
 $Q^\theta = \theta^\transpose  \psi$, a common choice of eligibility  
vector is defined through the $d$-dimensional recursion,
\begin{equation}
\elig_{n+1} = \lambda \disc \elig_n +\psi( Z_{n+1}), \ \  \elig_0 = \psi( Z_{0})  \, ,
\label{e:elig}
\end{equation}
where $ \lambda\in[0,1] $ is fixed, with $\lambda=0$ most common in applications. 

\label{e:Qintro}
\end{subequations}

The term \textit{Q-learning} is reserved for the case in which   $   \uQ^{\theta_n} (X_{n+1})  = \min_u   Q^{\theta_n} (X_{n+1},u)$, and the recursion \eqref{e:Qlambda} reduces to Watkins' algorithm  when using a tabular basis  \cite{watday92a,wat89}. 
See \cite{sutbar18,sze10,CSRL} for a range of interpretations of the algorithm.  
\textit{TD-learning} is the alternative in which  \eqref{e:Qintro} is constructed to approximate the fixed-policy Q-function.  In this case, there are three possible implementations: 
\begin{subequations}%
\begin{align}
  \uQ^{\theta_n} (X_{n+1})  & = \sum_u Q^{\theta_n}(x,u) \feex(u\mid x) &&  \textit{Natural}
  \\[-1em]
  & && \ x= X_{n+1}    \nonumber
  \\[1em]
  & =  Q^{\theta_n} (Z_{n+1})   &&  \text{On-policy}
   \label{e:onpolicy}
\\
& = Q^{\theta_n} (Z'_{n+1})   &&  \text{Split-sampling}
   \label{e:splitTD}
\end{align}
For the on-policy method    \eqref{e:onpolicy} it is assumed that the input $\bfmU$ used for training in \eqref{e:Qintro} is defined by the policy $\feex$.   
In the other two approaches the only restriction on the input is that it is adapted to the state process and results in a stable algorithm.  
   In the case of split sampling    \eqref{e:splitTD}
we define $Z'_{n+1} = (X_{n+1}, U'_{n+1})$ in which the random variable $U'_{n+1}$ is drawn randomly,
with distribution 
 $U'_{n+1} \sim \feex(\varble \mid x)$ when  $X_{n+1} =x$,
  independent of $\{U_{n+1}, Z_k : k\le n \}$.\label{e:uQintro}
\end{subequations}

\subsection{Convergence rates}

Any of the TD algorithms considered in this paper can be expressed in the form of a stochastic approximation (SA) recursion
\begin{equation}
\theta_{n+1}   =  \theta_n +  \alpha_{n+1}   f_{n+1}(\theta_n)  \,,  
\label{e:SA}
\end{equation}
with
 $  f_{n+1}(\theta_n)  = f(\theta_n, \Phi_{n+1})$ in which the definition of the stochastic process $\bfPhi$ depends on the specifics of the algorithm.  
 
 In our treatment of TD learning it is assumed that the policy used for training is \textit{oblivious} ($\feex$ does not depend on the parameter estimate).   
    It is further assumed that $\bfPhi$ is Markovian and has a unique steady state distribution $\uppi$,  which for TD learning requires that $\bfmZ$ has a unique steady state pmf, which is denoted $\occm$.  
    In on-policy TD(0) we may take $ \Phi_{n+1} = (Z_n, Z_{n+1})$.   For TD($\lambda$) with non-zero $\lambda$ we must include the eligibility vector, so that  $ \Phi_{n+1} = (Z_n, Z_{n+1}, \elig_{n+1})$ for each $n\ge 0$.

 \notes{Warning!    We cannot say that $\occm$ is the steady-state for $\bfPhi$ since it is a pmf on $\zstate=\xstate\times\ustate$.  Careful!!}

Potential limits of the SA recursion are  solutions to the root finding problem    $\barf(\theta^*) = 0$    in which 
$\barf(\theta) = \Expect_\uppi[ f(\theta,\Phi_{n+1})]$,  $\theta\in\Re^d$.  
 Convergence theory of SA is built around stability theory for  the  $d$-dimensional   mean flow $\ddt \odestate_t =  \barf(\odestate_t) $ introduced in \eqref{e:meanflowQ}.
The strongest conclusions for SA require that the mean flow is  exponentially asymptotically stable (EAS).

\notes{I removed \textit{
Under a randomized stationary policy} since that should go elsewhere
\\
Also, big error in definition of mean flow.   $\barf$ is the vector field, and the mean flow is an ODE}


It is well known that the rate of convergence of the mean square error is no faster than $O(1/n)$ except in exceptional settings such as quasi-stochastic approximation  \cite{laumey22c}.
When this rate is achieved we can typically also establish the following limit, which defines the \textit{asymptotic covariance}:
\begin{equation}
		\SigmaTheta \eqdef  \lim_{n\to\infty} {n} \Expect[  \tiltheta_n \tiltheta_n^\transpose ] \,, \quad
				\textit{where $\tiltheta_n =  \theta_n -\theta^*$ }
		\label{e:SigmaTheta}
\end{equation}
Under mild assumptions   we minimize the asymptotic covariance through choice of step-size followed by   averaging.  

Much of the theory of SA begins with the following assumptions on 
the step-size sequence: 
 \begin{equation}
\sum_{n=1}^\infty \alpha_n = \infty \,, \qquad
\sum_{n=1}^\infty \alpha_n^2 < \infty 
\label{e:stepass}
\end{equation}
In the discussion that follows and in numerical experiments we opt for the standard choice,
\begin{equation}
 \alpha_n = \min(\alpha_0,  n^{-\rho}) \,, \ \textit{ with $\alpha_0>0$ and $1/2<\rho <1$. } 
\label{e:alpharho}
\end{equation}
The upper bound on $\rho$ is imposed to   justify averaging,  the Polyak-Ruppert (PR) averaged estimates \cite{rup88,poljud92}, which are defined by 
\begin{equation}
	\thetaPR_N = \frac{1}{N-N_0+1} \sum_{k=N_0}^N  \theta_k\,,\qquad N\ge N_0
	\label{e:PR}
\end{equation}
in which the interval $\{0,\dots,N_0 \}$ is known as the \textit{burn-in} period.

Theory for additive white noise models is contained in \cite{poljud92} where it is shown that the resulting asymptotic covariance has the form 		\eqref{e:SigmaPRopt} 
in which $\Sigma_\Delta$ is the covariance of the additive noise.    It is known that the matrix $ \SigmaTheta^* $ is minimal in the matricial sense.  In the recent work \cite{borchedevkonmey25} it is shown that the same conclusions hold for general SA recursions with Markovian disturbance, with an alternative definition of the disturbance covariance:
\begin{equation}
\Sigma_\Delta  = \sum_{k=-\infty}^\infty R_\Delta(k)
\label{e:SigmaDelta}
\end{equation} 
where $R_\Delta(k)$ is the autocorrelation sequence for the zero mean sequence $\{   f(\theta^*, \Phi_{n})  :  n\in\intgr \}$    with $\bfPhi$   a stationary realization of the Markov chain.

\notes{I don't think this is needed --  Phi has been with us in nearly every paragraph: 
\\
associated with the stochastic approximation recursion \eqref{e:SA}.}

We also consider the asymptotic bias, defined by
\begin{equation}
	\upbeta_\uptheta
	\eqdef
	\lim_{n\to\infty}
	\frac{\Expect[\theta_n - \theta^*]}{\alpha_n}  \, ,
	\label{e:a-bias}
\end{equation} 
which under mild conditions admits the representation \eqref{e:biasformula}.

\subsection{Relative TD methods}

The expressions \eqref{e:SigmaPRopt} and  \eqref{e:biasformula} tell us that we can expect slow convergence when the condition number of $\barA$ is large.   
It is pointed out in \cite{devmey17b,devmey22} that in the case of tabular Q-learning, the matrix  $\barA$ is Hurwitz but has an eigenvalue that tends to zero as the discount factor tends to unity.    
Relative Q-learning was introduced in \cite{devmey22} to address this numerical instability and reduce the asymptotic covariance.  
Moreover, theory from \cite{devmey22} implies that   the greedy policy obtained from relative Q-learning is optimal in the tabular setting.

A relative TD or Q-learning 
algorithm  
  is defined by a recursion of the form \eqref{e:Qlambda} with a slight modification of the temporal difference sequence, 
\begin{equation}
\begin{aligned}
\Tdiff_{n+1}   &=  c(Z_n)   + \disc     \uQ^{\theta_n} (X_{n+1})  - Q^{\theta_n}(Z_n)   
\\
&\qquad -   \delrel \langle \upmu ,  Q^{\theta_n} \rangle \,  ,
\end{aligned}
\label{e:TD_theta_n_rel}
 \end{equation}
 in which $\upmu$ is a pmf on $\state\times \ustate$ (the \textit{baseline distribution})
 and $\delrel >0$. 

 It is shown under mild conditions on  $\upmu$  and  $\delrel >0$ that the resulting recursion is stable and the asymptotic covariance is uniformly bounded over $0\le \disc <1$:   Q-learning is treated in  \cite{devmey22}  and the simpler case of TD-learning is  treated in \cite{CSRL}.   However, to-date theory has been absent outside of the tabular setting. 
 
Theory of stability of TD learning starting with 
 \cite{tsivan97} requires consideration of the autocorrelation matrices   $R(k) \eqdef \Expect[\psi_{(0)}\psi_{(k)}^\transpose] $ for any integer $k$,  where $\psi_{(k)} = \psi(Z_k)$, and the expectation is taken in steady state.  The mean flow for the RTD($\lambda$) algorithm is given by $\barf(\theta) = \barA \theta + \barb$ with 
	\begin{align}
	\barA(\lambda;\delta_r) &= \Expect\left[\zeta_n\left[-\psi_{(n)} -\delrel \barpsi^\upmu + \disc \psi_{(n+1)} \right]^\transpose\right]\notag\\
	& = \barA(\lambda;0) - \frac{\delrel}{1-\lambda\disc} \barpsi[\barpsi^\upmu]^\transpose 
	\label{e:barA}
	\end{align}
in which $\barpsi = \Expect_\uppi[\psi_{(0)}]$, and $\barpsi^\upmu$ is the vector whose $i$th component is 
\begin{equation}
	\barpsi^{\upmu}_i = \langle \upmu, \psi_i \rangle, \qquad 1 \le i \le d.
\label{e:barpsiupmu}
\end{equation}

In \eqref{e:barA} and in the following development, the dependence of $\barA$ on $\lambda$ and $\delrel$ is made explicit when needed.

The second equality in \eqref{e:barA} follows from \eqref{e:elig} and stationarity. Taking expectation in \eqref{e:elig} gives
$
\Expect[\elig_n]
=
\lambda\disc \Expect[\elig_n] + \barpsi 
$,
and therefore
\[
\Expect[\elig_n]
=
\frac{1}{1-\lambda\disc}\barpsi \,,
\quad
\Expect\!\left[\elig_n(\barpsi^\upmu)^\transpose\right]
=
\frac{1}{1-\lambda\disc}\barpsi[\barpsi^\upmu]^\transpose \,,
\]
which implies the second equality in \eqref{e:barA}.


In the present paper we show that the attractive results obtained for tabular Q-learning or TD learning extend to TD learning with linear function approximation,  subject to assumptions on the pmf $\upmu$.   The strongest conclusions are obtained when this is taken to be the steady state pmf $\occm$ for $\bfmZ$.

   As this is not known a priori, we can substitute the empirical distribution   of the state–action process.    This reasoning leads to the following algorithm:

 \begin{subequations}%
 
\wham{\boldmath $\occm$-\textbf{Relative  TD($\lambda$) learning}}
 
For initialization $\theta_0\in\Re^d  $, the sequence of estimates are defined recursively
by \eqref{e:Qlambda} 
with eligibility vectors defined in \eqref{e:elig} and temporal difference sequence obtained through the following:
\begin{equation}
\begin{aligned}
\Tdiff_{n+1}   &=  c(Z_n)   + \disc     \uQ^{\theta_n} (X_{n+1})  - Q^{\theta_n}(Z_n)   
\\
&\qquad -   \delrel    \langle \occm_n ,  Q^{\theta_n} \rangle \,  ,
\end{aligned}
\label{e:TD_theta_n_rel2025}
 \end{equation}
 in which $ \langle    \occm_n ,  Q^{\theta_n} \rangle  =    \barpsi_n^\transpose \theta_n $ with
\begin{equation}
\barpsi_{n+1} 
  = \barpsi_n  +    \beta_{n+1} [ \psisub{n+1}   -  \barpsi_n ],\quad \barpsi_0=\psi(Z_0),
\label{e:barpsi}
\end{equation} 
and $\{ \beta_n \}$ is a vanishing step-size sequence also satisfying \eqref{e:stepass}.   For ease of exposition we assume a relatively large gain:  $\lim_{n\to\infty}  \beta_n/\alpha_n = \infty$. 
\notes{New RK: Should we explicitly define $\occm_n$?
\\
We really should!   I don't want to take up space on this for CDC.}

\label{e:occTD}
 \end{subequations}%

 \notes{The reader doesn't know about ``fixed on-policy'' algorithms, and the ``fixed'' isn't required here}
 
Theory of two time-scale SA justifies consideration of a   mean flow for approximation of the slow parameter sequence $\{\theta_n \}$ \cite{bor20a}.    It remains linear,   with  $\barA$   given by \eqref{e:barA} using $\upmu = \occm$:
\begin{equation}\label{e:barA_relative}
\barA(\lambda;\delta_r) = \barA(\lambda;0) - \frac{\delrel}{1-\lambda\disc} \barpsi\barpsi^\transpose.
\end{equation}
\notes{New RK: The bot says: "with adaptive (\occm_n), this is not strictly a single-time-scale linear flow; it is a limiting/two-time-scale statement." I assume this is OK with the assumption that $\beta_n/\alpha_n \to \infty$, but do we need to add a statement to clarify?
\\
Yes!  I made the change with reference to Borkar.}


\wham{Contributions}

While the algorithm \eqref{e:TD_theta_n_rel} was introduced more than five years ago, there has been no analysis outside of the tabular setting.
It turns out that a good choice for $\delrel$ and $\upmu$ is an art in general.  

Moreover, to-date there is not much theory providing sufficient conditions for stability of Q-learning outside of the tabular setting   \cite{mey24,mehmey26a}.   
 Consequently,  the theoretical results summarized here are restricted to RTD learning with linear function approximation.   We choose the on-policy variant based on    \eqref{e:onpolicy} since we are assured that the mean flow for TD learning is EAS under the assumptions of  \cite{tsivan97}.

The main new realization in this paper is that $\upmu = \occm$ is universally successful.   
For this choice of $\upmu $ and other choices, subject to assumptions, we obtain in general the following dichotomy:
\whamb   For the standard TD($\lambda$) learning algorithm the optimal asymptotic covariance  \eqref{e:SigmaPRopt}
and asymptotic bias 
	\eqref{e:a-bias}  may be unbounded as a function of discount factor $\disc\in (0,1)$.   
	
\whamb
Under mild assumptions the optimal asymptotic covariance and asymptotic bias for RTD learning are uniformly bounded over $\disc\in (0,1)$,  for any fixed $\lambda<1$.

\notes{I don't know what this means:  , together with favorable stability and conditioning properties.
\\
I think you mean better condition numbers for $\barA$ but the motivation here is not clear.
 }

  \notes{I'm not sure this adds much:
 Since this same matrix also appears in the formulas for the optimal asymptotic covariance and the asymptotic bias, the improved conditioning obtained through RTD learning is expected to have positive implications for finite-data error bounds as well as asymptotic variance and bias. }

\notes{New:  (revised by a bot!) The references to Moulines were critically important. He's done the best work. }
\wham{Literature}

 Relative Q-learning for average cost appeared in \cite{aboberbor01}. One decade later it was realized that the same technique could be used to obtain uniformly bounded rates of convergence in the discounted-cost setting, uniformly over $0<\disc<1$ \cite{devmey22}. These papers are entirely devoted to the tabular setting. Extensions of the algorithm to TD learning are contained in \cite{CSRL} (see Section 9.5.3 and the regenerative algorithm in Section 10.4.2). To the best of our knowledge, stability theory has been largely restricted to the tabular setting, with the exception of Theorem 10.15 of \cite{CSRL}, which treats the TD(1) algorithm for average cost. The motivation there was for applications to actor--critic methods.

The main results of this paper rest on establishing that the SA mean flow matrix $\barA$ is uniformly Hurwitz and bounded over the range of $(\lambda,\disc)$ considered. Bounds on the conditioning of this matrix are also required to obtain meaningful guarantees in finite-time analysis of TD learning and linear stochastic approximation. For instance, finite-time bounds for linear stochastic approximation and TD learning with constant step size are established in \cite{srikant2019}. More recent work considers TD learning with decreasing step sizes and provides rates of convergence for averaged iterates \cite{srikant2025}. Related finite-time high-probability bounds for Polyak--Ruppert averaged iterates of linear stochastic approximation are established in \cite{durmounausam24}.
In each of these works, the quantitative behavior depends explicitly on stability of the mean flow along with conditioning properties of the mean flow matrix $\barA$.

In \cite{srikant2019}, the obtained bounds are proportional to the condition number of the solution $P$ of the Lyapunov equation
$
\barA^\transpose P + P\barA = -I$,
which improves with improved conditioning of $\barA$. In \cite{srikant2025}, bounds are obtained on the Wasserstein distance between $\sqrt{N}(\thetaPR_N-\theta^*)$ and a zero-mean normal distribution with covariance matrix $\SigmaTheta^*$; these bounds likewise improve when $\barA$ is better conditioned. The results of \cite{durmounausam24}, following \cite{bacmou11}, similarly show that finite-time concentration bounds for averaged linear SA depend on the condition number of $\barA$.

The present work contributes to this line of research by showing that RTD learning, with an appropriately chosen baseline, yields a mean flow matrix $\barA$ that remains uniformly Hurwitz and well-conditioned as the discount factor approaches one. This leads to uniform bounds on both asymptotic covariance and bias, in contrast to standard TD methods where these quantities may diverge.

 \wham{Organization}
The remainder of the paper is organized as follows.
\Cref{s:main} presents the main theoretical results.
\Cref{s:sims} presents two simulation examples: one for a finite MDP in which all quantities (such as those defining asymptotic bias and covariance) are computable in close form, and a second example with continuous state and action space.   
Conclusions and directions for future research are contained in \Cref{s:conc},  and proofs of technical results are collected together in an appendix.

   
\section{Main Results}
\label{s:main}

The main results are restricted to on-policy TD learning (based on    \eqref{e:onpolicy}) in which $\bfmZ$ is a Markov chain on the finite state space $\zstate =\state\times\ustate$.   The $k$th  \textit{autocovariance}  matrix is denoted $\Sigma(k) \eqdef R(k) -\barpsi \barpsi^\transpose$,  where  the $k$th autocorrelation matrix was introduced earlier:    $R(k) \eqdef \Expect[\psi_{(0)}\psi_{(k)}^\transpose] $ with $\psi_{(k)} = \psi(Z_k)$.   
Assumption (A0) justifies these definitions and (A1) was introduced in \cite{tsivan97} in an early treatment of TD learning:   
\wham{(A0)}    $\bfmZ$ is a unichain and aperiodic Markov chain, with unique pmf $\occm$.  
\wham{(A1)}  $R(0) >0$  
	\ \  \textit{or}     \ \ 
 \textbf{(A1${}^{\bullet}$)}     $ \Sigma(0) >0$.

\notes{We never referred to (A2)  }
 
 Proofs of the main results surveyed this section may be found in the appendix.

\notes{New:  (we have to say what it is we are estimating!)}

\subsection{Bias}

The following result is a consequence of \eqref{e:barA} 
(which defines  $	\barA(\lambda;\delta_r) $) 
combined with the 
  Sherman--Morrison formula.     

\begin{proposition}[Bias]
\label[proposition]{t:biasBig}
Suppose that (A0) holds, and for fixed $\lambda\in [0,1]$, suppose that the matrices $\barA_0 \eqdef \barA(\lambda;0) $  
 and  $  \barA(\lambda;\delrel) $    	are invertible. 
 Let  $\theta^*(0)$, $\theta^*(\delrel)$ denote the respective stationary points for the mean flow.   
  Then,
\[
\theta^*(\delrel)  
=
[I + G]
\theta^*(0) 
\,,
\qquad 
G =   
\frac{\delrel}{1-\lambda\disc}\,
\frac{\barA_0^{-1}\barpsi[\barpsi^\upmu]^\transpose  }{1-[\barpsi^\upmu]^\transpose \barA_0^{-1}\barpsi}    
\]
\qed
  \end{proposition}%

Hence the estimate obtained from RTD-learning may be far from what is obtained using the standard algorithm.     
In special cases we find that the difference $Q^{\theta^*(0)} (z)  - Q^{\theta^*(\delrel)} (z)  $ is independent of $z$,
so that the corresponding $ Q^{\theta^*(\delrel) }$-greedy policies do not depend on $\delrel$
  \cite{devmey22}.


We next provide justification for
  a finite (and in general non-zero) limit
  defining the asymptotic bias in \eqref{e:a-bias}.    The following proposition follows from the general theory in  \cite{laumey22c}.

 \begin{subequations}%

\begin{proposition}[Asymptotic bias]
\label[proposition]{t:bias}
Consider the SA recursion \eqref{e:SA} with $ f_{n+1}(\theta) = A(\Phi_{n+1}) \theta + b(\Phi_{n+1})$ in which $\bfPhi$ is unichain and aperiodic,
and the steady-state mean $\barA = \Expect[ A(\Phi)]$ is Hurwitz. 
Suppose the step-size is of the form \eqref{e:alpharho}.      Then, the limit \eqref{e:a-bias} exists and has the form  \eqref{e:biasformula}
in which 
$
\barUpupsilon
=
\Expect[\Upupsilon_{n+1}^*]
$
with expectation in steady-state, and 
\begin{equation}
\Upupsilon_{n+1}^*
=
\big(
A_{n+1} - \haA_{n+1}
\big)
\big(
A_{n+1}\theta^* + b_{n+1}
\big),
\end{equation}
where $A_{n+1} \eqdef A(\Phi_{n+1})$, $b_{n+1} \eqdef b(\Phi_{n+1})$, and $\haA_{n+1} \eqdef \haA(\Phi_{n+1})$ is a zero-mean $d\times d$ stochastic process defined through the Poisson equation  $
\Expect[ \haA_{n+1}  \mid \Phi_0^n] = \haA_n  - A_n  +\barA$.
\qed
\end{proposition}

\label{e:biasComponents}
 \end{subequations}%

The proposition only applies to TD(0) learning since when $\lambda>0$ we must take  $ \Phi_{n+1} = (Z_n, Z_{n+1}, \elig_{n+1})$, which violates the specific geometric ergodicity assumption imposed in \cite{laumey22c}.   We conjecture that a similar result holds for general $\lambda$.

\subsection{Stability and asymptotic variance}


We have noted that  TD-learning algorithms may exhibit very slow convergence when the discount factor is close to unity, and one explanation is that the Jacobian $\barA$ may be ill conditioned.    Alternatively,  one finds that the estimation problem itself is fundamentally ill-conditioned.     Consider for illustration the fixed policy Q-function  \eqref{e:Qfee}.
Under mild conditions one obtains the approximation, 
\begin{equation}
Q(x,u) =  H(x,u) + \frac{\eta}{1-\disc}  + o(1-\disc)
\label{e:QandPoisson}
\end{equation}
where $o(1-\disc) \to 0$ as $\disc\uparrow 1$,  
  $\eta $ is the average cost,  and $H$ a solution to Poisson's equation,
$\eta + H(x,u) = c(x,u) +    \Expect [ H(Z_1) \mid   Z_0    = (x,u ) ]   $  \cite[Chapter~6]{CSRL}.


   If the basis is chosen to approximate $Q$ then from \eqref{e:QandPoisson}
    it is not unreasonable to assume there is a vector $\xi \in\Re^d$ satisfying
\begin{equation}
\xi ^\transpose \psi(z) = 1 \,, \quad  \textit{for $ z\in \state\times\ustate$ satisfying $\occm(z)>0$}
\label{e:v1}
\end{equation}
In the case of a tabular basis we obtain \eqref{e:v1} using $\xi _i = 1$ for each $i$.   

Recall that  $\barpsi^{\upmu}\in\Re^d$ is defined in \eqref{e:barpsiupmu}.

\begin{theorem}[RTD($\lambda$) may be unstable]
\label[theorem]{t:unstable}
Suppose that  \emph{(A0)-(A1)} hold and there is a solution $\xi \in\Re^d$ to \eqref{e:v1}.   
Suppose that
    $\upmu$ is the pmf  on $\state\times \ustate$ used in the RTD($\lambda$) learning algorithm.     
      Then,   

\whamrm{(i)}   If     $\xi^\transpose \barpsi^{\upmu} >0$ then for each $\lambda \in [0,1)$  there exists  
$\delrel^0>0$ such that
the mean flow for 
RTD($\lambda$) learning is EAS for any 
 $\disc \in [  0,1] $       
 and $\delrel \in (0,\delrel^0)$.
 
    \whamrm{(ii)}   If $\xi^\transpose \barpsi^{\upmu} < 0$ then for each $\lambda \in [0,1)$   there exists  
$\delrel^0>0$ and  $\disc^0>0$ 
such that $\barA(\lambda;\delrel)$ contains an eigenvalue in the strict right half plane  
whenever $\disc \in [ \disc^0, 1]$ and $\delrel \in (0,\delrel^0)$. 
\qed
 \end{theorem}

 \notes{April: I cleaned up the theorem (need to be careful with lambda), and for CDC added a proof sketch which I know many reviewers would ask for}

To ensure stability we might design the basis so that $\xi^\transpose \psi(z) > 0$ for each $z$.   We opt for the following alternative:
  To ensure stability we assume
$ \Sigma(0) >0$  (i.e.,  (A1${}^{\bullet}$) holds).

Observe that subject to   \eqref{e:v1} we have $\xi ^\transpose \psisub{k} =  \xi ^\transpose \barpsi =1$ for all $k$, and hence
\[
\xi ^\transpose \Sigma(0) \xi = \xi ^\transpose \Expect[ \psisub{k} \psisub{k}^\transpose] \xi -  (\xi ^\transpose \barpsi )^2  =0  
\]
Hence (A1${}^{\bullet}$)    fails.  
However, the existence of $\xi$ is well motivated only when we wish to directly estimate the $Q$ function using a standard TD learning algorithm.

\begin{theorem}[Stability and Variance]
\label[theorem]{t:var}

 Under~\emph{(A0)-(A1${}^{\bullet}$)},  for any $\epsy \in (0,1)$,   

\whamrm{(i)}
 There exists  $\delrel^0>0$ such that $\barA(\lambda;\delrel)$ obtained from RTD($\lambda$) learning is Hurwitz for any 
 $\lambda, \disc \in [  0,1 ] $ satisfying $\disc \lambda \le 1-\epsy$, 
  and $\delrel \in [0,\delrel^0)$.    Moreover, the condition number and optimal asymptotic covariance are uniformly bounded:
 \begin{equation}
\sup \, \big[ \cond(\barA(\lambda;\delrel)) + \trace(\SigmaTheta^*)   \big]
  < \infty
\label{e:conparA}
\end{equation} 
 where the supremum is over this range of  $\lambda,\disc , \delrel$.
 
\whamrm{(ii)}  
If $\upmu = \occm$ then  
 the mean flow associated with the RTD($\lambda$) algorithm is stable for any  $\delrel\ge 0$,
and any 
 $\disc , \lambda\in [0,1]$  satisfying $\disc \lambda<1$.     
 Moreover,  \eqref{e:conparA}  holds with the supremum over all such $\disc,\lambda$
 satisfying $\disc \lambda \le 1-\epsy$ and  $\delrel \in [  0,1 ] $.
\qed
\end{theorem}

\subsection{Sensitivity}

The next result considers sensitivity of variance and bias of
$\occm$-RTD($\lambda$) learning for small $\delrel$.   
We restrict to $\lambda=0$, and recall that 
 in this case $\delrel=0$ corresponds to the standard TD(0)-learning algorithm.

\Cref{t:var}~(ii) extends to this algorithm, since the joint process $\{  (\theta_n; \barpsi_n  ):  n\ge 0\} $ is the output of a two time-scale linear SA recursion in which the mean flow associated with $\{\theta_n \}$  is identical to that considered in \Cref{t:var}~(ii).    It is also convenient to consider an alternative to simplify analysis.
 
 \wham{\boldmath  Fixed $\occm$-\textbf{Relative  TD($0$) learning}}
The on-policy version of the algorithm is expressed  as
\begin{equation}
\begin{aligned}
\theta_{n+1} & = \theta_n + \alpha_{n+1}   [A_{n+1} \theta_n + b_{n+1}  ]   
\\
&A_{n+1} =  \psisub{n} \big[ -\psisub{n} +\disc \psisub{n+1} \big]^\transpose -\delrel  \barpsi\barpsi^\transpose
\\
&  b_{n+1} = \csub{n}\psisub{n}\, \ \ \end{aligned}
\label{e:occ-rel-det}
\end{equation}
The vector $\barpsi$ is computed a priori,  or estimated   through Monte-Carlo \eqref{e:barpsi}
with $\beta_n=1/n$.  
\notes{YOu can't really `compute' using Monte-Carlo}

\notes{Error commented out here.   The mean flow doesn't change for the fixed version}
 
We denote  by  $\theta^*(\delrel)$ the stationary point for the mean flow,   $\SigmaTheta^*(\delrel)$ the asymptotic covariance matrix  \eqref{e:SigmaPRopt}  and 
$ \upbeta_\uptheta(\delrel)$ the value of \eqref{e:a-bias}  for arbitrary $\delrel$ in RTD-learning.
In the following results, we also require explicit indication of dependency on $\delrel$ for auxiliary variables such as  $\barUpupsilon(\delrel)$.   
When $\delrel =0$ we drop the dependency to simplify the resulting equations. Since the following results only apply to the $\lambda=0$ case, we also drop this dependency from $\barA(\lambda;\delrel)$.

 \begin{subequations}%

\begin{theorem}[Sensitivity]
\label[theorem]{t:sens}
Consider the  $\occm$-RTD(0) learning as defined in \eqref{e:occTD} subject to 
\emph{(A0)-(A1${}^{\bullet}$)}.
The algorithm is convergent for any value of $\delrel\ge 0$ and we have the following sensitivity formulae for asymptotic covariance and bias: 
 \begin{eqnarray}
	\begin{aligned}
		\frac{d}{d\delrel}  \SigmaTheta^*(\delrel)    \big|_{\delrel=0}   & = 
		\barA^{-1}\Sigma_\Delta'\barA^{-\transpose}
		\\
		&\quad 
		+
		\barA^{-1} \barpsi \barpsi^\transpose \SigmaTheta^* 
		  + [  \barA^{-1}\barpsi \barpsi^\transpose  \SigmaTheta^*  ] ^{\transpose}
	\end{aligned}
	\\ 
	\frac{d}{d\delrel} 
	\upbeta_\uptheta (\delrel) \big|_{\delrel=0}  =
	\frac{1}{1-\rho}
	\barA^{-1}
	\Big[
 	\barpsi \barpsi^\transpose  \upbeta_\uptheta
	+
	\barUpupsilon'
	\Big] 
	\label{eq:bias-derivative-appendix}
\end{eqnarray}
The matrix   $\Sigma_\Delta'\eqdef \frac{d}{d\delrel}  \Sigma_\Delta(\delrel)\Big|_{\delrel=0}$ 
and vector $\barUpupsilon'\eqdef \frac{d}{d\delrel} \barUpupsilon(\delrel)\Big|_{\delrel=0}$
are each identically zero when using the variant \eqref{e:occ-rel-det}.
\qed
\end{theorem}

 \end{subequations}%

The theorem combined with  \eqref{e:SigmaPRopt} implies that a bound on    $\|  		\frac{d}{d\delrel}  \SigmaTheta^*(\delrel)   \|$  can be expected to grow as fast as $\| 	\barA^{-1} \|^3$ and \eqref{eq:bias-derivative-appendix} combined with
 \eqref{e:biasformula} tell us that a bound on    $\| \frac{d}{d\delrel} 	\upbeta_\uptheta (\delrel) \|$ will grow as  $\| 	\barA^{-1} \|^2$.
An example of the steep rate of change in bias is illustrated in  \Cref{f:biasCostsensitivity} in our first numerical example.

\section{Simulations}
\label{s:sims} 	

 We survey results from two examples to  
 illustrate the main results of the paper.   The first considers a finite-state, finite-action MDP, while the second examines a stochastic speed-scaling model with a continuous state space. 
 In both cases, comparisons between TD and  $\occm$-RTD illustrate the improved stability and variance properties of the proposed method.

\notes{I'd prefer to say we are illustrating, not evaluating
\\
evaluate the}

 \notes{I shrunk all the figs for arXiv.   One has to be returned to wrapped}

\subsection{Finite-state Finite-action MDP}


\begin{figure}[h]
  \centering
  
  	\includegraphics[width= 0.7\hsize]{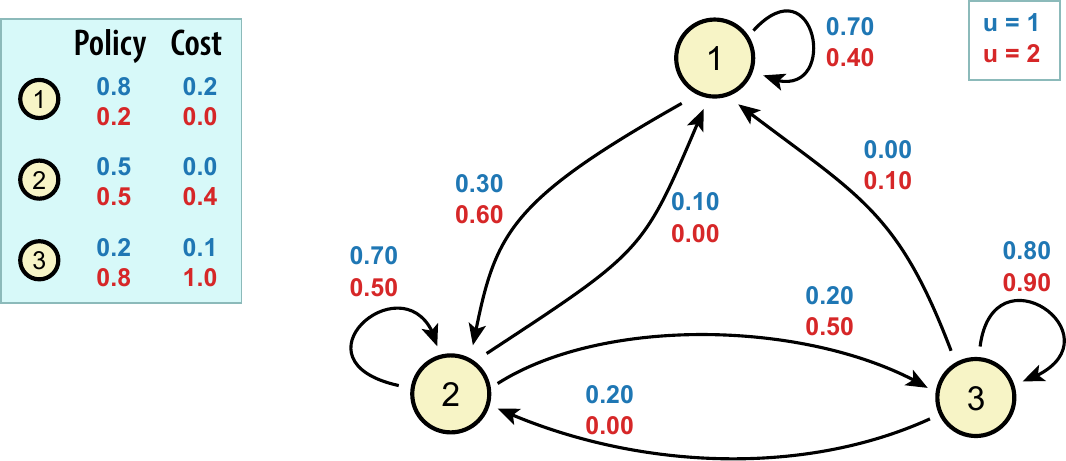}  

  \caption{3-state 2-action MDP model, with randomized policy.  Variants of TD learning were performed with discount factor $\disc=0.99$,  and step-size   \eqref{e:alpharho} with $
\rho    = 0.65$  and $\alpha_0      = 2$.   }
    \label{f:3X2U}
\end{figure}
 
 \Cref{f:3X2U} shows an MDP example with state space $\state=\{1,2,3\}$ and action space $\ustate=\{1, 2\}$, along with the randomized policy considered in experiments.   For example,  $\feex(1\mid 1) =\Prob\{ U_k = 1 \mid   X_k =1 \} = 0.7$  and $\feex(1\mid 3) =\Prob\{ U_k = 1 \mid   X_k =3 \}  =  0.7$.

The optimal average cost policy is unique and deterministic with ${\fee^*}(1) = 2$ and ${\fee^*}(2) = {\fee^*}(3) =1$.    Under this policy   the transition matrix and one-stage cost vector are
\[
P_{\fee^*} =
\begin{bmatrix}
0.40 & 0.60 & 0.00\\
0.10 & 0.70 & 0.20\\
0.00 & 0.20 & 0.70
\end{bmatrix},
\qquad
c_{\fee^*} =
\begin{bmatrix}
0\\
0\\
0.1
\end{bmatrix},
\]
  the invariant pmf is  $
\uppi_{\fee^*} = [1, 6,6]/13$ and   average cost  $\eta^* = \uppi_{\fee^*}(c_{\fee^*}) = 3/65$.

 Since the optimal policy for the average-cost problem is unique,
it follows that $\fee^*$ is also optimal for the $\disc$-discounted problem for all
$\disc$ sufficiently close to~1.

\begin{figure}[h]
  \centering
  
  	\includegraphics[width= 0.7\hsize]{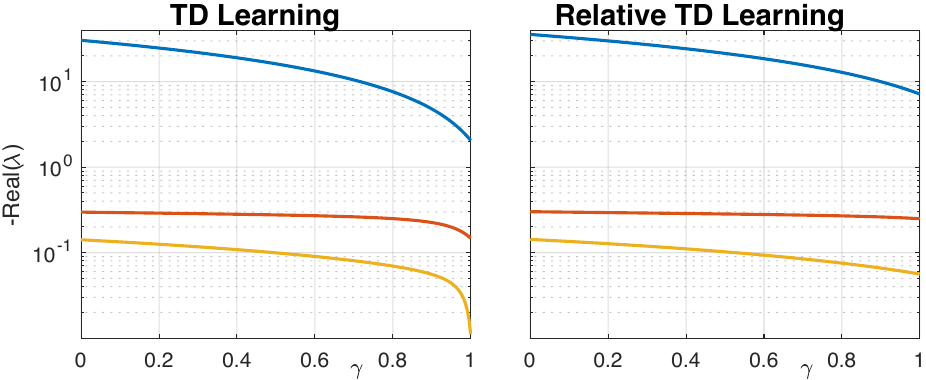}

  \caption{Eigenvalues as a function of discount factor $\disc$.   One eigenvalue approaches zero for the algorithm using $\delrel=0$.}
  \label{f:EigComparison}
\end{figure}

%
%
%

%
%
%
%


For the randomized policy shown in the figure, the induced transition matrix and one-stage cost   are 
\[
P_{\feex}=
\begin{bmatrix}
0.64 & 0.36 & 0\\
0.05 & 0.60 & 0.35\\
0.08 & 0.04 & 0.78
\end{bmatrix},
\qquad
c_{\feex}=\begin{bmatrix}
0.16\\
0.20\\
0.72
\end{bmatrix}
\]
The invariant pmf   is $\uppi_{\feex} = [85, 108, 315]/508$,
and hence the average cost is far greater than $\eta^*$:  
\[
\eta^{\feex}= \uppi_{\feex}(c_{\feex}) 
=
\frac{587}{1016}
\approx 0.578.
\]

%

The 3-dimensional basis vector was chosen to be $	\psi(x,u) 
	=
	[x ; \; u;\; xu]$, which satisfies $\Sigma(0) >0$.

 The numerical results that follow were based on $M=200$ independent runs of length $N=10^6$.   For histograms we increased the number of samples to $\nsnap \times M$ by selecting $\nsnap = 5$ parameter estimates from the final 80\%\ of each run.   These were chosen approximately uniformly on the natural time scale associated with SA approximation theory, 
resulting in  
\begin{equation}
		N_i =
		\biggl(
		N_1^{1-\rho}
		+
		\frac{i-1}{\nsnap-1}
		\bigl(
		N_{\nsnap}^{1-\rho} - N_1^{1-\rho}
		\bigr)
		\biggr)^{\frac{1}{1-\rho}} \,, \ \ 1\le i\le \nsnap \, ,
\label{e:nsnap}
\end{equation}
with $N_{\nsnap}=N$ and $N_1 = 0.8 N$.
Explanation is provided in the Appendix.

\begin{figure}[h]
  \centering
  
  	\includegraphics[width= 0.75\hsize]{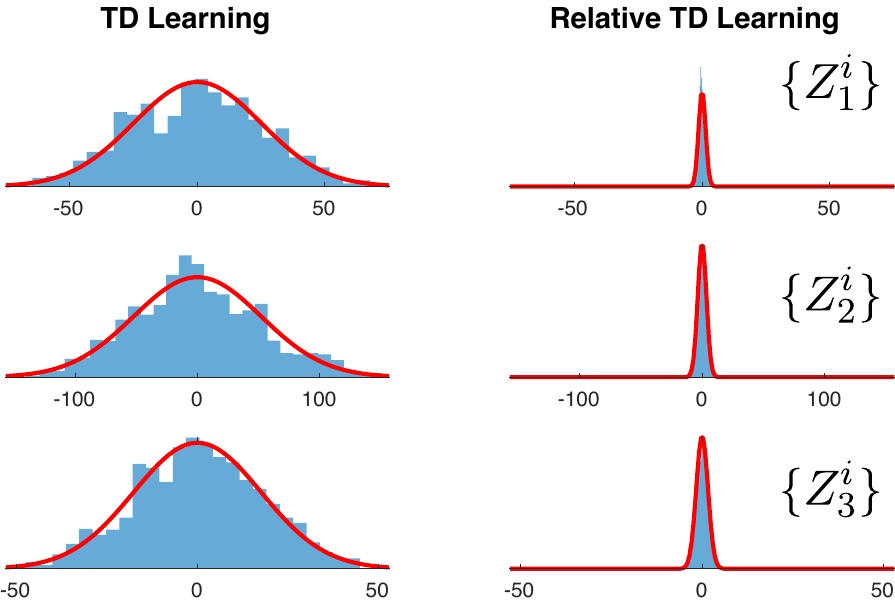}

\caption{Histograms of the scaled and centered parameter estimates for TD(0) and $\occm$-RTD(0) in the finite state-action model using Polyak--Ruppert averaging. The red solid curves show the corresponding theoretical Gaussian approximations predicted by \eqref{e:SigmaPRopt} and $\{\mathcal{Z}_j^{i,m}\}$ denote the samples used to form the histogram of the $j$-th component.}
  \label{f:hists}
\end{figure}

It is known that PR averaging not only achieves the optimal asymptotic covariance, but also ensures that the CLT holds \cite{borchedevkonmey25}.  To illustrate this conclusion we construct a histogram of the centered and scaled errors:
For each $1\le m\le M$ and $1\le i\le  \nsnap$ (recall  
\eqref{e:nsnap}),    denote by  $ \thetaPR_{i,m}$ the value of 	\eqref{e:PR}  with the value of $N$ replaced by $N_i > N_0$ in the $m$th independent run.   
The empirical mean is given by
\begin{equation}
\barthetaPR 
	=
	\frac{1}{M\nsnap}
	\sum_{m=1}^{M}\sum_{ i=1}^{\nsnap}  \thetaPR_{i,m}  \,, 
\end{equation}
Consider the centered and scaled error,
\begin{equation}
	\clZ^{i,m}
	=
	\sqrt{ N_i - N_0}
	\bigl( 
	 \thetaPR_{i,m}  - \barthetaPR 
	\bigr) \,, 
\end{equation}
in which \(1 \le m \le M\), and  $1 \le i \le \nsnap$.  The  three histograms obtained from the three components of $\{ 	\clZ^{i,m} \} \subset \Re^3$ are shown in  \Cref{f:hists}.   Each appears to be consistent with a Gaussian random vector  with covariance \eqref{e:SigmaPRopt} (which was computed exactly based on model parameters).
       This accuracy is less tight with time horizons   below $N=10^5$.

\begin{figure}[h]
 	\centering     
 	\includegraphics[width= 0.45\hsize]{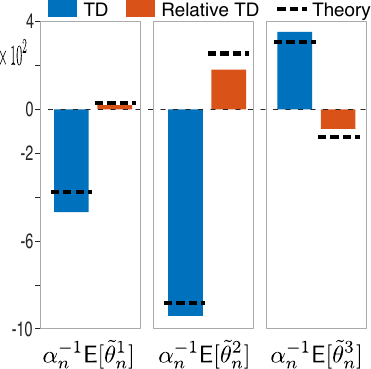}  
  \caption{Bias realized vs theory.}  
   \label{f:biasThy}
\end{figure}

%

 \Cref{f:EigComparison} shows the eigenvalues of $\barA$ as a function of the discount factor. For the standard TD(0) algorithm, the matrix $\barA$ becomes close to singular as $\disc\uparrow 1$, resulting in poor conditioning. The \(\occm\)-RTD(0) algorithm avoids this behavior.

	   \Cref{f:biasThy}  shows the components of the empirical mean of $\{  [\theta_n^i - \theta^*]/\alpha_n   : 1\le i\le M \}$ with $n=10^6$ and $M=200$ using $\occm$-RTD learning with  
	     $\delrel = 1/2$.   The theoretical value is  obtained from the asymptotic bias formula \eqref{e:biasformula}.

	     \bigskip

\begin{figure}[h]
  \centering
  
  	\includegraphics[width= 0.8\hsize]{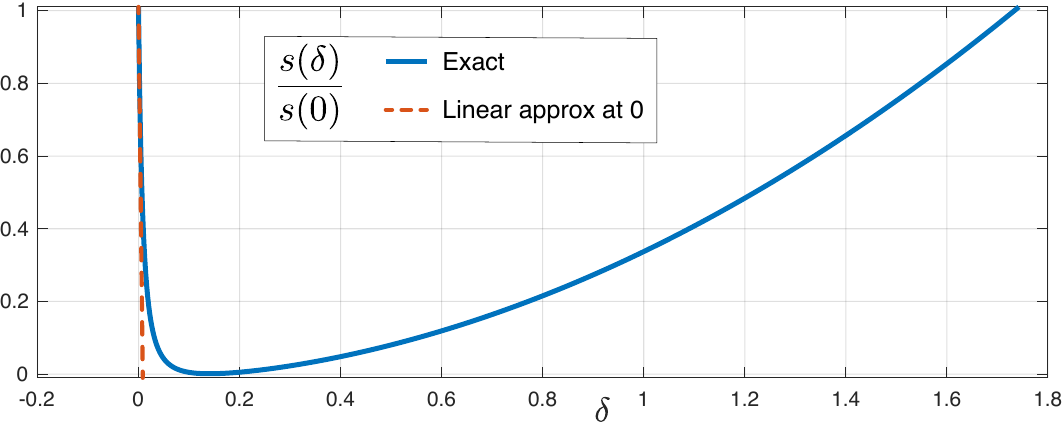}

  \caption{Squared-norm of normalized bias as a function of $\delrel$.   The red dashed line is the predicted slope }
    \label{f:biasCostsensitivity}

\end{figure}

The plot of   $s(\delta) = \|\upbeta_\uptheta (\delta) \| ^2$ shown in 
    \Cref{f:biasCostsensitivity}  is also obtained from \eqref{e:biasformula}:
    The red dashed line has slope $s'(0) = 2 \upbeta_\uptheta^\transpose \upbeta_\uptheta'$ where $ \upbeta_\uptheta'$  is given in \eqref{eq:bias-derivative-appendix}.

 \clearpage


\notes{
\rd{RK: All of the theory in the paper has assumed finite state and action spaces, so technically does not apply to this example, should we say something about that?} 
\\
SM:  Done.   BTW, is ``benchmark'' appropriate here?  I don't know, but I'm in a rush so I'll leave it.  }

\subsection{Speed scaling}

The second   example is the speed-scaling model, which serves as a benchmark for studying reinforcement learning with continuous state and action spaces. The state space is $\state=\Re_+$ and the state-dependent action space is $\ustate(x)=[0,x]$.  The dynamics are given by
\begin{equation}
X_{k+1} = X_k - U_k + A_{k+1}
\end{equation}
where $\{A_k\}$ is i.i.d.\ with finite mean and variance. 
The state $X_k$ represents the workload in a queue, and the control $U_k$ denotes the service rate.
See \cite[Sec.~7.4]{CSRL} for history.

\begin{figure}[h]
  \centering
  \includegraphics[width=0.7\hsize]{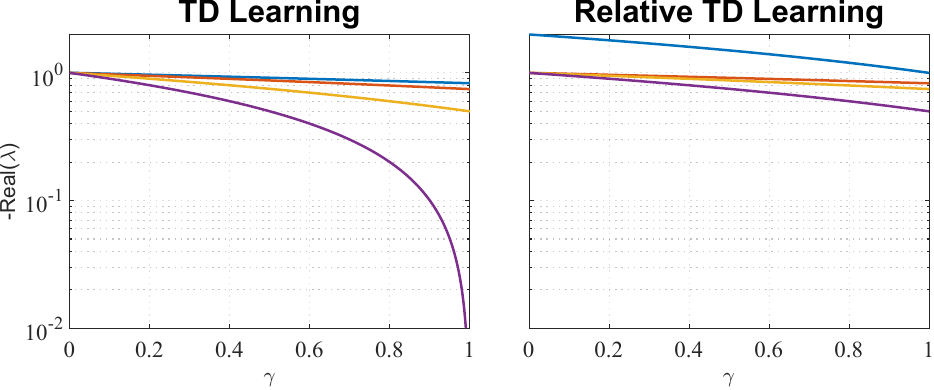}
  \caption{Eigenvalues as a function of discount factor $\disc$ for the speed-scaling model.}
  \label{f:EigCompar_Speed}
\end{figure}

The one-stage cost is selected to be
\begin{equation}
c(x,u) = x + \frac{r}{2}u^2 \,, 
\label{e:sscost}
\end{equation}
where \(r>0\) is a design parameter. This cost captures the trade-off between delay and power consumption: increasing the service rate reduces the queue length but incurs higher instantaneous cost.

While the main results in this paper have been established only for finite state and action models, most results easily extend to this general setting since all of the SA theory cited has been established for a general state space Markov chain $\bfPhi$ in the recursion \eqref{e:SA}.

Our goal is to estimate the fixed policy Q-function with policy   $u(x)=\kappa x$ with $0<\kappa <1$;   The value $\kappa = 0.5$ was used in numerical experiments,   $\{A_k\}$ was chosen to be i.i.d.\ with a Gamma distribution having mean $5$ and variance $10$, and $r=10$ was used to define the cost function \eqref{e:sscost}.

\notes{Big changes:}
We adopt a linear function approximation architecture to approximate the state-action value function using the 4-dimensional basis  
$\psi(x,u)
	=
	[c(x,u) ; \, x^{3/2} ; \, -(1+\sqrt{x})u ;\, 1- {u}/{\sqrt{x+1}}]$.   \notes{Note Matlab notation using semicolons for vectors}  
The basis is motivated by an approximation of $Q^\star $,  devised through two steps.  First,  \eqref{e:QandPoisson} admits the
extension  $Q^\star(x,u) =  H^\star(x,u) +  \eta^*/(1-\disc)   + o(1-\disc)$ in which $\eta^*$ is the optimal average cost, and $H^\star$ is the relative value function solving the   average cost optimality equations.     Approximation theory based on a fluid model reveals that $H^\star$ grows as $x^{3/2}$ for large $x>0$.     
For the linear policy under consideration, the fixed policy Q-function grows as $x^2$ rather than $x^{3/2}$;   however,    we do not expect to obtain a tight approximation uniformly over $\zstate $,  so we opt for a basis similar to what has been used in prior work.

Finally, we chose $\disc=0.9$, and for \(\occm\)-RTD(0)  we set $\delta_r=1$.
 The step-size sequence was selected according to \eqref{e:alpharho} with $\alpha_0=0.2$ and $\rho=0.6$.

	 \notes{I don't think a reader will understand this, so I shortened:
	 \\
	  structure of the speed-scaling model and the large-state approximation of the value function, following the discussion in \cite[Section~7.4]{CSRL}. The term $x^{3/2}$ is suggested by the fluid-model approximation, while the terms involving $u$ are chosen so that the induced greedy policy retains the form $a+b\sqrt{x}$, which is natural for this model. We do not include an independent constant basis function, since this can lead to rank deficiency in the autocorrelation matrix and a nearly singular TD mean matrix as $\disc \uparrow 1$. Instead, the constant offset is absorbed into the fourth basis through $1-u/\sqrt{x+1}$. This ensures that $\psi(0,0)\neq 0$, so the approximation retains a bias term, while avoiding the numerical difficulties associated with a stand-alone constant feature.
	}


The main results are   illustrated here for the case where $\lambda=0$.  For RTD, the baseline distribution is taken to be the empirical distribution (\(\occm\)-RTD(0) learning).   Polyak--Ruppert averaging is applied in both cases for variance reduction,  with run-length $N=10^{7}$ and $20\%$ of the trajectory treated as burn-in, so that $N_{\mathrm{burn}}=0.2N$.  To estimate asymptotic covariance and bias we ran $M=100$ independent simulations.  For histograms we increased the number of samples to $100\times M$ by selecting $\nsnap = 100$ parameter estimates from the final $20\%$ of each run (recall 	\eqref{e:nsnap}).
 \notes{April 1:   I brought back in 	\eqref{e:nsnap} for continuity with the first example.   We need to discuss the chose of $\nsnap =100$--this is far too big since successive samples are so highly correlated.  I chose the value $\nsnap=5$ for reasons you can find in my notes.  This is fine for now, but let's please discuss.    }

\Cref{f:EigCompar_Speed} shows the eigenvalues of $\barA$ as a function of the discount factor $\disc$ for both TD and RTD in the speed-scaling model,  showing that the condition number of $\barA$ is uniformly bounded over $0\le \disc\le 1$ for RTD(0).

\begin{figure}[h]
  \centering
  \includegraphics[width=0.7\hsize]{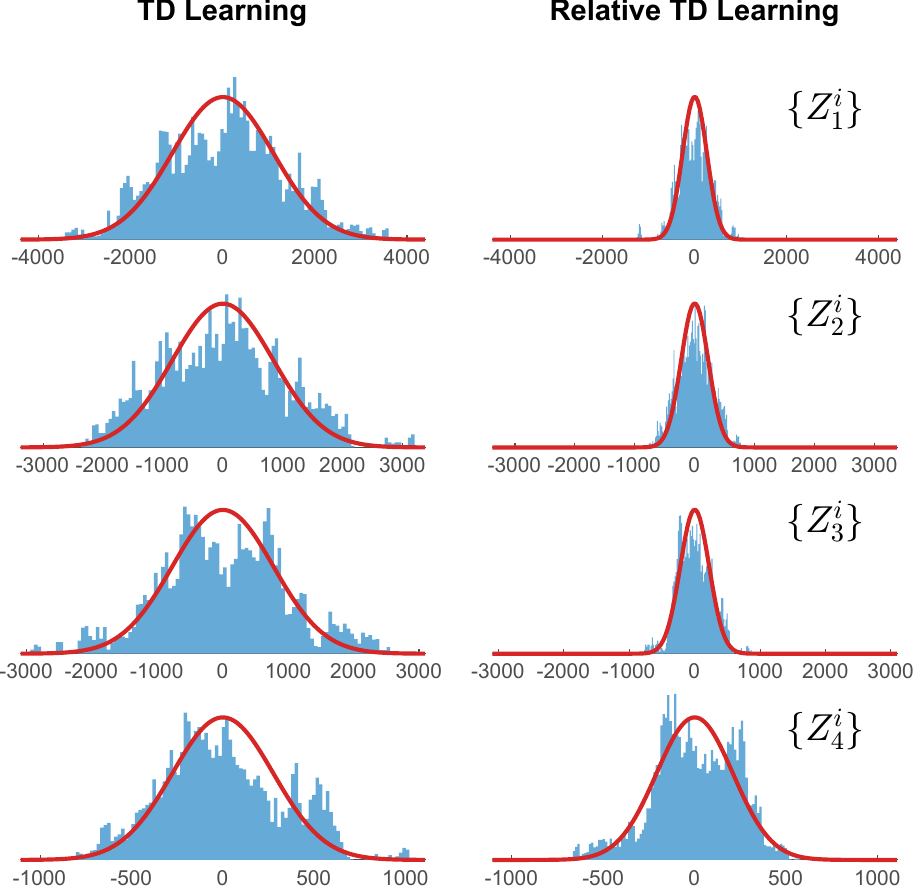}
\caption{Histograms of the scaled and centered parameter estimates for TD(0) and $\occm$-RTD(0) in the speed-scaling model using Polyak--Ruppert averaging. The red solid curves show the corresponding theoretical Gaussian approximations predicted by \eqref{e:SigmaPRopt} and $\{\mathcal{Z}_j^{i,m}\}$ denote the samples used to form the histogram of the $j$-th component.}
  \label{f:histsSpeed}
\end{figure}

 \Cref{f:histsSpeed} illustrates that  the  Polyak--Ruppert averaged parameter estimates exhibit behavior consistent with the Gaussian limit predicted by the CLT, for both TD(0) and $\occm$-RTD(0).    

\subsection{Discussion.}
The simulation results in both examples are consistent with the theoretical analysis of stability, covariance, and bias developed in \Cref{s:main}. In particular, \Cref{t:var} predicts that the \(\occm\)-RTD(0) algorithm maintains a Hurwitz matrix $\barA$ with uniformly bounded condition number. The uniform bound for \(\occm\)-RTD(0) is illustrated in  \Cref{f:EigComparison,f:EigCompar_Speed}, for both the finite-state and speed-scaling examples. In both examples, the standard TD(0) algorithm exhibits an eigenvalue approaching zero as $\disc \uparrow 1$, while the \(\occm\)-RTD(0) algorithm maintains eigenvalues bounded away from the right half plane in $\Co$, as predicted by \Cref{t:var}.

\notes{Error here:
indicating potential instability
\\
Stability holds}

\notes{Also, note that what is crucial is Hurwitz, so I rephrased.  Previously you had this: 
\\
maintains eigenvalues bounded away from zero}
 
\Cref{f:hists,f:histsSpeed} show that the asymptotic covariance of \(\occm\)-RTD(0) is reasonably well captured by the Gaussian approximation based on \eqref{e:SigmaPRopt}.
The variance reduction for \(\occm\)-RTD(0), compared to TD(0), as observed in \Cref{f:hists,f:histsSpeed}, is attributed to the larger condition number of $\barA$ under TD(0) (see \Cref{f:EigComparison}).

The bias results in the finite-state example are also consistent with theory. As observed in \Cref{f:biasThy}, the empirical normalized bias closely matches the theoretical prediction in    \Cref{t:bias}. Moreover, as observed in \Cref{f:biasCostsensitivity}, the slope of the squared bias at $\delta_r=0$ is consistent with the sensitivity analysis established in \Cref{t:sens}.

\section{Conclusions}
\label{s:conc} 	

This paper introduces a relative TD learning algorithm based on an empirical baseline distribution. In addition to stability and uniform bounds on the asymptotic covariance, it is also demonstrated that this choice of baseline distribution leads to improved conditioning of the mean dynamics, uniformly over the discount factor, and results in significantly reduced variance compared to standard TD learning. In addition, we also establish sufficient conditions for instability of the mean flow. The theoretical results are supported by simulation studies in both finite-state-action MDPs and MDPs formulated on general state and action spaces.

Future work will focus on extensions to policy iteration and  Q-learning. 

\notes{Error here:  control algorithms such as  Q-learning.   Few would say that Q-learning is a control algorithm}

\notes{I do not agree with this statement.   Let's please discuss:
Another important direction is the derivation of finite-time error bounds for the empirical relative TD algorithm.
}

  \clearpage

\bibliographystyle{abbrv}
\def\cprime{$'$}\def\cprime{$'$}

  \clearpage

\appendix

\section{Appendices}

The mean flow $\barA \eqdef \barA(\lambda; 0)$ for the standard on-policy TD($\lambda$) learning algorithm   is linear, with  
\begin{equation}
\begin{aligned}
 \barA  &= - \sum_{k=0}^\infty  (\lambda\disc)^k R_{k}   +  \disc  \sum_{k=0}^\infty   (\lambda\disc)^k  R(k+1)  
\\
&=
-  R(0) +   (1- \lambda  )  \disc \sum_{k=0}^\infty (\lambda\disc)^k  R(k+1)  
\end{aligned}
\label{e:barA_TDlambda}
\end{equation}
It was first established in  \cite{tsivan97} that $\barA$ is Hurwitz when $R(0)$ is full rank.    
      The conclusion that is commonly applied is
\begin{equation}
\tmptheta^\transpose  \barA   \tmptheta  \le -(1-\disc)  \tmptheta^\transpose R(0)\tmptheta  \,, \quad \tmptheta\in\Re^d
\label{e:bvrHurwitz}
\end{equation}
whose proof follows from Cauchy-Schwarz inequality, giving $\tmptheta^\transpose R(k)\tmptheta \le \tmptheta^\transpose R(0)\tmptheta$ for any $k$.

In some cases, such as the tabular setting, the bound is nearly tight as $\disc\uparrow 1$,  which leads to the instability result \Cref{t:unstable}~(ii).  

\subsection{Dirichlet forms and eigenvalues of $\barA$}

See \cite{levperwil17,fil91}
for background on spectral gaps and Dirichlet forms for  
Markov chains.     This theory is set in a Hilbert space setting, in which a transition matrix is regarded as a linear operator on $L^2 = L^2(\uppi)$.  

\notes{liuols21 useful?}  

Introduce two scalars $\beta = \lambda\disc$  and $\varrho = \disc (1 -\lambda)/(1-\beta) <1$, 
so that \eqref{e:barA_TDlambda} becomes
\begin{equation}
\begin{aligned}
\barA &=  -(1- \varrho) R(0)  -  \varrho   M_\beta  
\\
\textit{with} \ \ 
M_\beta &= R(0) -  (1-\beta)\sum_{k=0}^\infty \beta^k R(k+1)
\end{aligned}
\label{e:MA}
\end{equation}
In the special case $\lambda=0$ we obtain $M_\beta = R(0) - \disc R(1)$.  
To improve \eqref{e:bvrHurwitz} we obtain a lower bound on $\tmptheta^\transpose M_\beta \tmptheta$ for arbitrary $\beta \in  [0, 1)$.

\notes{  \rd{RK: I do not think we have defined the notation $\Expect_{\feex}$. Does the above equation serve as the definition?}
\\
That's a ChatGPT goof.   Let's stick to uppi}

 For each $\tmptheta\in\Re^d$ we associate a function $g_\tmptheta\colon\zstate\to \Re$ via  $g_\tmptheta(z)=\tmptheta^\transpose\psi(z)$ for $z\in  \zstate = \state\times\ustate$, 
 defined so that  $\sum_z \uppi(z) g_\tmptheta(z)  = \Expect_\uppi [g_\tmptheta(Z)]$.

 Consider the transition matrix
\[
K_\beta \eqdef (1-\beta)\sum_{k=0}^\infty \beta^k P^{k+1}  
\]
 Using the operator theoretic notation $K_\beta g \big|_z = \sum_j K_\beta(z,z') g(z')$  for functions $g\colon\zstate\to\Re$,  we obtain,
\begin{equation}
\tmptheta^\transpose  M_\beta \tmptheta    
=    \langle  g_\tmptheta,  [I- K_\beta] g_\tmptheta \rangle_{L^2} 
\eqdef \sum_z  \uppi(z) g_\tmptheta (z)  \{  [I- K_\beta] g_\tmptheta \}\big|_z
\label{e:Dirchelet1}
\end{equation}

A lower bound  is obtained from standard
spectral-gap inequalities for finite irreducible Markov chains. 

For a transition matrix $T$ with unique invariant pmf $\uppi$, its  
$L^2(\uppi)$ adjoint is expressed
 $T^*(z,z') = \uppi(z') T(z',z)/\uppi(z)$.
Letting   
 $\bfPsi$ denote a stationary version of the Markov chain with this transition matrix,     the associated 
 \textit{Dirichlet form} is defined for functions $g,h\colon\zstate\to\zstate$ by 
\[
\clE_T(g,h) \eqdef 
\half\Expect_\uppi [(g(\Psi_1)-h(\Psi_0))^2]  
\]
When $g=h$ this may be expressed,
\begin{equation}
\clE_T(g,g)  =  \langle  g,  ( I- T)g \rangle_{L^2} =  \langle  g,  ( I- T^*)g \rangle_{L^2} 
\label{e:DirichletAdj}
\end{equation}  

\begin{lemma}
\label[lemma]{t:Poincare}
 Let $T$ denote an arbitrary transition matrix on $\zstate$ with unique invariant pmf $\uppi$,  with spectral gap $\upgamma$.    If $T$ is reversible then the Poincar\'e inequality holds:   for any $g\colon\zstate\to\Re$,
 \[
 \clE_T(g,g)  \ge \upgamma  \|\tilg\|_{L^2(\uppi)}^2   \,, \qquad \textit{ where $\tilg = g - \uppi(g)$.  
}
 \]
\end{lemma}

Application of the lemma is made possible through consideration of the   ``reversibilization''
$S_\beta=(K_\beta+K_\beta^*)/2$, where $K_\beta^*$ is the 
$L^2(\uppi)$ adjoint of $K_\beta$.   The identities in \eqref{e:DirichletAdj} imply that  
    $\clE_{K_\beta} (g,g) = \clE_{S_\beta}(g,g)$, allowing us to apply \Cref{t:Poincare} to obtain   a lower bound on $\tmptheta^\transpose M_\beta \tmptheta$ that is independent of $\beta$.      
    
 Let $\upgamma_\beta>0$ denote the spectral gap of $S_\beta$.

 \begin{subequations}%

\begin{proposition}
\label[proposition]{t:Dirichlet}
Suppose that $\bfmZ$ is unichain and aperiodic.   Then,

\whamrm{(i)}  For $\tmptheta\in\Re^d$,
\begin{equation}
 \tmptheta^\transpose  M_\beta \tmptheta   \ge \upgamma_\beta  \, \tmptheta^\transpose  \Sigma(0) \tmptheta 
\label{e:Mbdd}
\end{equation}

\whamrm{(ii)} 
 $\bar\upgamma_P = \inf \{ \upgamma_\beta  :  0\le \beta <1 \}$ is strictly positive.

\whamrm{(iii)} For any non-negative pair $\lambda, \disc$ statisfying $\lambda \disc <1$ the matrix $\barA$ in \eqref{e:MA}
satisfies the bound 
\begin{equation}
 \tmptheta^\transpose  \barA  \tmptheta   \le -  \bar\upgamma_P \tmptheta^\transpose  \Sigma(0) \tmptheta \,, \quad \tmptheta\in\Re^d
\label{e:Abdd}
\end{equation}
\end{proposition}

 \end{subequations}%

 \wham{Proof}
 Observe that $ \upgamma_\beta $ is continuous as a function of $\beta$, with $\upgamma_\beta >0$ for $0\le \beta < 1$  and  $\upgamma_\beta(\beta) \uparrow 1$ as $\beta\uparrow 1$.     
 Conclusion (ii) thus follows, i.e.,
  $ \bar\upgamma_P >0 $.

As for (iii),  the bound \eqref{e:Mbdd} combined with \eqref{e:MA}   imply \eqref{e:Abdd}:
 \[
  \tmptheta^\transpose  \barA  \tmptheta   
=
   -(1- \varrho)   \tmptheta^\transpose R(0) \tmptheta  -  \varrho    \tmptheta^\transpose M_\beta  \tmptheta   \le -  \bar\upgamma_P \tmptheta^\transpose  \Sigma(0) \tmptheta 
\]
where the second inequality uses $R(0) \ge \Sigma(0)$.  
Hence the proof is complete on establishing (i).

Since $S_\beta$ is reversible with respect to $\uppi$, \Cref{t:Poincare}  
implies
\[
\langle g,(I-S_\beta)g\rangle_{\feex}
\ge \upgamma_\beta \|\tilg\|_{L^2(\uppi)}^2 
\]
Applying this with $ g_\tmptheta$ gives the desired bound in (i):
\[
\tmptheta^\transpose M_\beta \tmptheta
\ge
\upgamma_\beta \|\tilg_\tmptheta\|_{L^2(\uppi)}^2
=
\upgamma_\beta\, \tmptheta^\transpose\Sigma(0) \tmptheta.
\]
\qed 

\subsection{Proof of \Cref{t:unstable}}

Subject to the assumptions of \Cref{t:unstable}, 
we will show shortly that  when $\disc=1$ the matrix $\barA$ defined in \eqref{e:MA}  has a simple eigenvalue at the origin,
and all others in the strict left half plane.     To prove the theorem we consider conditions under which this eigenvalue moves to the left or right left half plane for RTD($\lambda$) learning for small $\delrel>0$.

\begin{lemma}  
\label[lemma]{t:lambdaflow}
Let $\Xi\in\Re^{d\times d}$ have a simple eigenvalue at $0$,
with eigenvector $\eta \neq 0$ such that
$
\Xi \eta  =  
\eta ^\transpose \Xi = 0 $,  normalized to $\|\eta\|=1$.

Let $v,w\in\Re^d$ be nonzero and denote $\Xi_t \eqdef \Xi + t\, v w^\transpose$.   Then there is $\epsy>0$ and a differentiable function
$\{\kappa_t : 0\le t\le \epsy \}$ such that $\kappa_0 =0$ and $\kappa_t$ is an eigenvalue of $\Xi_t$ for $0\le t\le \epsy$.   
Its derivative at the origin may be expressed,   
\[
\kappa'_0 =  (\eta ^\transpose v)(w^\transpose \eta )
\]
\end{lemma}

\wham{Proof}
An application of the implicit function theorem implies the existence of $\epsy>0$ such that there is an eigenvalue-eigenvector pair
$(\lambda_t,\eta^t)$ for $\Xi_t$ for $0\le t\le \epsy$:
\begin{equation}\label{eq:eigpair}
\Xi_t \eta _t = \kappa_t\,\eta _t \,, 
\end{equation}
  with initialization $\lambda_0 =0$ and $\eta^0 = \eta$, 
and with $\| \eta^t \| =1$ for each $t$.

Differentiate \eqref{eq:eigpair} at $t=0$, use $\kappa_0=0$ and apply the product rule to obtain
\[
(v w^\transpose)\eta  + \Xi \eta '_0 = \kappa'_0\,\eta  + \kappa_0\,\eta '_0
= \kappa'_0\,\eta ,
\]
Left-multiply by $\eta ^\transpose$ to obtain
\[
\eta ^\transpose (v w^\transpose) \eta  + \eta ^\transpose \Xi \eta '_0 = \kappa'_0\, \eta ^\transpose \eta .
\]
Using $\eta ^\transpose \Xi=0$ and $\eta ^\transpose \eta =1$ gives
$
\kappa'_0 = \eta ^\transpose (v w^\transpose) \eta  = (\eta ^\transpose v)(w^\transpose \eta )$.
\qed

We have  $\barA(\lambda;\delrel) = \barA(\lambda;0) - \frac{\delrel}{1-\lambda\disc} \barpsi [\barpsi^\upmu]^\transpose$
for RTD($\lambda$) learning, 
with $\barA \eqdef \barA(\lambda;0)$ defined in \eqref{e:MA}.   This expression invites the application of \Cref{t:lambdaflow} with appropriate choice of $\Xi$ together with 
 $t =  {\delrel}/(1-\lambda\disc)$, 
$ v =  -   \barpsi $,  and $w = \barpsi^\upmu$.

Under the assumption that $R(0)>0$ it follows that the null space of $\Sigma(0)$ coincides with the linear span of vectors satisfying  \eqref{e:v1}.    This null space has dimension no more than one:
	 if $\xi^1,  \xi^2$ are linearly independent solutions to  \eqref{e:v1}, then their difference is in the null space of $R(0)$.    

Based on this structure we may apply 
\Cref{t:lambdaflow} with $\Xi$ equal to $\barA$ with $\disc =1 $.    The bound   \eqref{e:Abdd} implies that $\Xi$ 
has all eigenvalues in the strict left half plane, except a simple eigenvalue at zero with eigenvector $\xi$.

It follows from \Cref{t:lambdaflow} that there exists $\delrel^0>0$ and a differentiable eigenvalue branch $\kappa(\delrel)$ of $\barA(\lambda;\delrel)$ for $\delrel \in [0,\delrel^0)$ satisfying $\kappa(0)=0$, with derivative
\[
\kappa_0'
=
(\xi^\transpose v)(w^\transpose \xi)
=
- (\xi^\transpose \barpsi)(\xi^\transpose \barpsi^{\upmu}).
\]
 
 Since $\xi$ satisfies \eqref{e:v1}, we have $\xi^\transpose \barpsi =1 > 0$, and hence the sign of $\kappa_0'$ is determined by $\xi^\transpose \barpsi^{\upmu}$.

\whamrm{(i)} If $\xi^\transpose \barpsi^{\upmu} > 0$, then $\kappa_0' < 0$, and hence for $\delrel > 0$ sufficiently small the eigenvalue $\kappa(\delrel)$ lies in the strict left half plane. Since all other eigenvalues of $\barA$ are already in the strict left half plane, it follows by continuity that $\barA(\lambda;\delrel)$ is Hurwitz for all $\delrel \in (0,\delrel^0)$ and $\disc$ sufficiently close to $1$. 

\notes{Not clear, so kill for now:
The extension to $\disc \in [0,1)$ follows from continuity of $\barA$ in $\disc$.
}

\whamrm{(ii)} If $\xi^\transpose \barpsi^{\upmu} < 0$, then $\kappa_0' > 0$, and hence for $\delrel > 0$ sufficiently small the eigenvalue $\kappa(\delrel)$ lies in the strict right half plane. Since the eigenvalue at zero persists for $\disc$ sufficiently close to $1$, this implies that there exists $\disc^0 < 1$ and $\delrel^0 > 0$ such that $\barA(\lambda;\delrel)$ has an eigenvalue in the strict right half plane for all $\disc \in [\disc^0,1)$ and $\delrel \in (0,\delrel^0)$.
\qed
 \notes{No need to say "completes proof" before qed}

\subsection{Proof of \Cref{t:var}}

\Cref{t:Dirichlet} tells us that $\barA(\lambda;0)$ obtained from   TD($\lambda$) learning is Hurwitz for any discount factor, including $\disc = 1$, provided $\Sigma(0)>0$.   Part (i) immediately follows.

To establish~(ii), apply \Cref{t:Dirichlet} to obtain $\bar\upgamma_P>0$ such that
\[
\tmptheta^\transpose \barA(\lambda;0) \tmptheta
\le
-\bar\upgamma_P \tmptheta^\transpose \Sigma(0) \tmptheta ,
\qquad \tmptheta\in\Re^d ,
\]
uniformly over all $\disc,\lambda \in [0,1]$ satisfying $\disc\lambda<1$.
If $\upmu=\occm$, then
\[
\barA(\lambda;\delrel)
=
\barA(\lambda;0)
-
\frac{\delrel}{1-\lambda\disc}\,\barpsi\barpsi^\transpose .
\]
Hence,
\[
\begin{aligned}
	\tmptheta^\transpose \barA(\lambda;\delrel) \tmptheta
	&=
	\tmptheta^\transpose \barA(\lambda;0) \tmptheta
	-
	\frac{\delrel}{1-\lambda\disc}(\tmptheta^\transpose \barpsi)^2
	\\
	&\le
	-\bar\upgamma_P \tmptheta^\transpose \Sigma(0) \tmptheta
	-
	\frac{\delrel}{1-\lambda\disc}(\tmptheta^\transpose \barpsi)^2 .
\end{aligned}
\]
Since $\Sigma(0)>0$, there exists $\epsy_0>0$ such that
\[
\tmptheta^\transpose \Sigma(0) \tmptheta \ge \epsy_0 \|\tmptheta\|^2 ,
\qquad \tmptheta\in\Re^d .
\]
Since $\delrel\ge 0$, $\disc,\lambda\in[0,1]$, and $\disc\lambda<1$, we have
\[
-\frac{\delrel}{1-\lambda\disc}(\tmptheta^\transpose \barpsi)^2 \le 0 .
\]
Therefore,
\[
\begin{aligned}
	\tmptheta^\transpose \barA(\lambda;\delrel) \tmptheta
	&\le
	-\bar\upgamma_P \epsy_0 \|\tmptheta\|^2
	-
	\frac{\delrel}{1-\lambda\disc}(\tmptheta^\transpose \barpsi)^2
	\\
	&\le
	-\bar\upgamma_P \epsy_0 \|\tmptheta\|^2 ,
	\qquad \tmptheta\in\Re^d .
\end{aligned}
\]
uniformly over all $\delrel\ge 0$ and all $\disc,\lambda \in [0,1]$ satisfying $\disc\lambda<1$.
It follows that $\barA(\lambda;\delrel)$ is Hurwitz for every $\delrel\ge 0$.
Moreover, the preceding bound implies that the smallest singular value of $\barA(\lambda;\delrel)$ is bounded away from zero uniformly over this parameter range, and hence $\barA(\lambda;\delrel)^{-1}$ is uniformly bounded. Since $\barA(\lambda;\delrel)$ is also uniformly bounded, it follows that
\[
\sup \cond\bigl(\barA(\lambda;\delrel)\bigr) < \infty .
\]
Finally, since
\[
\SigmaTheta^*
=
\barA(\lambda;\delrel)^{-1}
\Sigma_\Delta
\barA(\lambda;\delrel)^{-\transpose},
\]
the uniform bounds on $\barA(\lambda;\delrel)^{-1}$ and $\Sigma_\Delta$ imply
\[
\sup \trace(\SigmaTheta^*) < \infty .
\]
Consequently,
\[
\sup \, \big[ \cond(\barA(\lambda;\delrel)) + \trace(\SigmaTheta^*) \big] < \infty ,
\]
where the supremum is over all $\delrel\ge 0$ and all $\disc,\lambda \in [0,1]$ satisfying $\disc\lambda<1$.
This proves~(ii).
\qed


\subsection{Proof of \Cref{t:sens}}

	\notes{Delete for CDC:}
We start with a technical results on the sensitivity of $\theta^*(\delrel)$ and $\barA(\delrel)^{-1}$ to changes in $\delrel$ at $\delrel=0$. These results are used in the proof of \Cref{t:sens}.
\begin{lemma}
\label[lemma]{t:A_sens}
Provided $\barA(0)$ is full rank, we have
\[
\begin{aligned}
	\frac{d}{d\delrel}  \theta^*(\delrel) \big|_{\delrel=0} 
	&=  \barA^{-1}  \barA' \theta^*
	\\
	\barA' 
	\eqdef
	\frac{d}{d\delrel}  \barA(\delrel) \big|_{\delrel=0}
	&=  - \barpsi \barpsi^\transpose 
	\\
	(\barA^{-1})'
	\eqdef
	\frac{d}{d\delrel}  \barA(\delrel)^{-1} \big|_{\delrel=0}
	&=  \barphi \barphi^\transpose\, ,
	\quad
	\textit{with $\barphi = \barA^{-1} \barpsi$}
\end{aligned}
\]
\end{lemma}

 \wham{Proof}
The derivative formula for $\barA$ follows from the identity  
$ \barA(\delrel) =  \barA  - \delrel \barpsi\barpsi^\transpose$.

Since $\theta^*(\delrel)$ is the equilibrium of the mean flow, we have
\[
\barA(\delrel)\theta^*(\delrel)=\barb,
\]
with $\barb$ independent of $\delrel$. Differentiating with respect to
$\delrel$ gives
\[
\barA'\theta^* + \barA
\frac{d}{d\delrel}\theta^*(\delrel)\Big|_{\delrel=0}
= 0.
\]
Hence
\[
\frac{d}{d\delrel}\theta^*(\delrel)\Big|_{\delrel=0}
=
-\barA^{-1}\barA'\theta^*.
\]

The final identity is the quotient rule for differentiation of a matrix inverse: 
 \[
(\barA^{-1})'
=
\barA^{-1}\barpsi \barpsi^\transpose \barA^{-1}
=
\barphi \barphi^\transpose,
\quad
\textit{with $\barphi = \barA^{-1}\barpsi$.}
\]
 \qed


 	From \eqref{e:SigmaPRopt}, for each $\delrel$,
	\[
	\SigmaTheta^*(\delrel)
	=
	\barA(\delrel)^{-1}\,
	\Sigma_\Delta(\delrel)\,
	\barA(\delrel)^{-\transpose}.
	\]
	Differentiating with respect to $\delrel$ and applying the product rule,
	\begin{align*}
		\frac{d}{d\delrel}\SigmaTheta^*(\delrel)
		&=
		(\barA^{-1})'\,\Sigma_\Delta\,\barA^{-\transpose}
		+
		\barA^{-1}\,\Sigma_\Delta'\,\barA^{-\transpose}
		\\
		&\quad
		+
		\barA^{-1}\,\Sigma_\Delta\,(\barA^{-\transpose})'.
	\end{align*}
	Using the identities
	\[
	(\barA^{-1})' = -\barA^{-1}\barA'\barA^{-1},
	\qquad
	(\barA^{-\transpose})'
	= -\barA^{-\transpose}(\barA')^{\transpose}\barA^{-\transpose},
	\]
	we obtain
	\begin{align*}
		\frac{d}{d\delrel}\SigmaTheta^*(\delrel)
		&=
		-\barA^{-1}\barA'\barA^{-1}\Sigma_\Delta\barA^{-\transpose}
		+
		\barA^{-1}\Sigma_\Delta'\barA^{-\transpose}
		\\
		&\quad
		-\barA^{-1}\Sigma_\Delta\barA^{-\transpose}
		(\barA')^{\transpose}\barA^{-\transpose}.
	\end{align*}
Substituting \eqref{e:SigmaPRopt} into the first and third terms yields
\begin{align*}
	\frac{d}{d\delrel}\SigmaTheta^*(\delrel)\Big|_{\delrel=0}
	&=
	\barA^{-1}\Sigma_\Delta'\barA^{-\transpose}
	\\
	&\quad
	-\barA^{-1}\barA'\SigmaTheta^*
	-
	\bigl[\barA^{-1}\barA'\SigmaTheta^*\bigr]^{\transpose}.
\end{align*}
  
To obtain the bias sensitivity formula, recall from \eqref{e:biasformula} that
\[
\upbeta_\uptheta
=
\frac{1}{1-\rho}\,
\barA^{-1}\barUpupsilon .
\]
	Differentiating yields
	\begin{align*}
		\frac{d}{d\delrel}\upbeta_\uptheta(\delrel)
		&=
		\frac{1}{1-\rho}
		\Big[
		(\barA^{-1})'\barUpupsilon
		+
		\barA^{-1}\barUpupsilon'
		\Big].
	\end{align*}
	Substituting $(\barA^{-1})' = -\barA^{-1}\barA'\barA^{-1}$ and using
	$\barA^{-1}\barUpupsilon = (1-\rho)\upbeta_\uptheta$, we obtain
	\[
	\frac{d}{d\delrel}\upbeta_\uptheta(\delrel)\Big|_{\delrel=0}
	=
	\frac{1}{1-\rho}\,
	\barA^{-1}
	\bigl[
	-\barA'\upbeta_\uptheta
	+
	\barUpupsilon'
	\bigr]
	\]
\qed

\subsection{Sub-sampling for histograms}
\notes{Delete for CDC:}
The motivation for \eqref{e:nsnap} comes from the ODE/SDE approximation for stochastic approximation. 

An SA recursion  can be interpreted as a noisy Euler approximation of the mean flow \eqref{e:meanflowQ}, for which the \textit{natural time scale} is defined by the partial sums,
\[
	\tau(n) = \sum_{k=1}^n \alpha_k .
\]
The ODE approximations of SA theory are expressed, 
\[
	\theta_n   \approx \odestate^{(k)}_{\tau(n)} \,, \quad n\ge k\ge 0\,,  
\]
in which $\{  \odestate^{(k)}_t  :  t\ge \tau(k) \}$ denotes the solution of the mean flow \eqref{e:meanflowQ},
 initialized at $ \odestate^{(k)}_{\tau(k)} = \theta_k$.    For vanishing stepsize, the approximation becomes increasingly tight as $k\to \infty$   \cite{bor20a}.

	For step size $\alpha_n = c n^{-\rho}$, we obtain the approximation
	\begin{equation}
		\tau(n) \approx \frac{c}{1-\rho} n^{1-\rho} .
	\end{equation}
	Hence, if snapshots are chosen uniformly on the natural time scale over the interval $[N_1,N_{\nsnap}]$, then for $1 \le i \le \nsnap$ the corresponding iteration indices are approximately given by \eqref{e:nsnap}.   Approximate uniformity is imposed in an attempt  to  minimize the dependency between samples for a single run.

%
%
%
  \end{document}